\documentclass[runningheads]{llncs}
\usepackage[T1]{fontenc}
\usepackage{graphicx}
\usepackage{subcaption}
\usepackage{tikz}
\usepackage{xcolor}
\usepackage{algorithm}
\usepackage{algorithmic}

\begin{document}
\title{Identifying High-Confidence Social Biases in LLMs for Trustworthy Conversational Tutoring Agents}
\titlerunning{Accepted for AIED 2026}
\author{Aitor Arronte Alvarez\inst{1}\orcidID{0000-0002-1605-1772} \and \\
Naiyi Xie Fincham \inst{1}\orcidID{0000-0002-4959-436X}}
%Third Author\inst{3}\orcidID{2222--3333-4444-5555}}
%
\authorrunning{A. Arronte Alvarez and N. X. Fincham}
% First names are abbreviated in the running head.
% If there are more than two authors, 'et al.' is used.
%
\institute{University of Hawaii at Manoa, Honolulu HI 96822, USA 
%Springer Heidelberg, Tiergartenstr. 17, 69121 Heidelberg, Germany
%\email{lncs@springer.com}\\
%\url{http://www.springer.com/gp/computer-science/lncs} \and
%ABC Institute, Rupert-Karls-University Heidelberg, Heidelberg, Germany\\
\email{\{arronte,naiyixf\}@hawaii.edu}}
\maketitle              % typeset the header of the contribution
\begin{abstract}
Conversational tutoring agents have been shown to improve learning engagement and student outcomes, and large language models (LLMs) are increasingly used in these systems to provide scalable, personalized feedback. However, LLMs may perpetuate or amplify stereotypical social biases, posing particular risks in educational settings.
In this study, we evaluate LLMs in conversational tutoring scenarios to identify high-confidence social biases, instances where models are unable to identify biased judgments in tutoring conversations with strong confidence that may affect their reasoning and the feedback they provide to learners. We present a new dataset generation method that enables bias evaluation under naturalistic instructional conditions by regenerating student–AI tutor interactions and introducing turns with controlled bias derived from a benchmark dataset.
Using this data, we assess multiple LLMs’ ability to detect stereotypical biases and analyze the confidence and reasoning underlying their responses through computational and human evaluations. We find that bias detection is substantially more challenging in conversational tutoring contexts than in benchmark-based evaluations, and that state-of-the-art LLMs are overconfident in their incorrect assessments of stereotypical bias statements. Moreover, model confidence strongly influences reasoning and feedback, highlighting the risks of overconfidence biased behavior in LLM-based tutoring agents. We conclude by discussing implications, mitigation considerations, and directions for future research.
\end{abstract}
\section{Introduction}

Conversational tutoring systems have been associated with improved learning efficiency and increased cognitive engagement \cite{graesser2001intelligent,rus2013recent}. Recent advances in transformer based Large Language Models (LLMs) have overcome many prior limitations of educational technologies \cite{kasneci2023chatgpt}, enabling personalized learning experiences at scale \cite{park2024empowering,weber2021pedagogical}. In particular, LLM-based conversational tutoring agents support adaptive, context-aware interactions that respond to individual students’ needs and provide dialogic reasoning, feedback, and scaffolding, and are perceived by learners as supportive and helpful \cite{labadze2023role,fincham2024using,schmucker2024ruffle}. However, LLMs still produce hallucinations \cite{huang2025survey}, exhibit significant social biases \cite{gallegos2024bias}, amplify cognitive biases \cite{cheung2025large}, and tend to be overconfident in their judgments \cite{pawitan2025}, which is especially problematic in educational settings where highly confident reasoning but incorrect or biased judgments can mislead learners.

While substantial attention has been given to factual evaluations of LLMs, including some work in educational applications \cite{estevez2025evaluation}, large-scale evaluation of AI-powered educational systems remains an emerging area with significant limitations \cite{maurya2025unifying}. In particular, the evaluation of social biases in educational contexts have been studied in terms of their technical \cite{lee2024life} and social \cite{weissburg2025llms,warr2024implicit} implications and possible safeguards in a range of subject domains, but the confidence and uncertainty associated with model judgments and the risks they pose in education remains an underexplored area. As LLM-based tutoring agents become more widely adopted, evaluations must be conducted within their specific educational contexts rather than relying solely on general-purpose benchmarks. However, collecting and evaluating educational data presents additional challenges, including ethical and privacy considerations \cite{huang2023ethics}.

In this paper, we address this gap by examining how the internal confidence of state-of-the-art LLMs affects their reasoning and the feedback they provide to students during the identification of stereotypical bias in student–tutor conversations, particularly in cases of incorrect and overconfident judgments. Our study focuses on the context of second language (L2) acquisition, where LLMs capabilities of supporting personalized interactive learning have prompted substantial scholarly attention and practical development. We first propose a method to create student-tutor dialogue datasets for stereotypical bias identification. The method deidentifies authentic student–tutor language-learning conversations and fully regenerates each dialogue using an LLM (DeepSeek), preserving the linguistic characteristics and instructional structure of the original interaction while introducing a controlled stereotypical turn derived from the StereoSet benchmark dataset \cite{nadeem-etal-2021-stereoset}. Second, we evaluate multiple LLMs on the resulting dataset to investigate the following driving research question: \textbf{How do state-of-the-art LLMs detect stereotypical bias in conversational tutoring contexts when compared to benchmark datasets, and how does their confidence, especially when incorrect, affect their reasoning and feedback provided to students?} We address this motivating question through a combination of computational and human evaluations. The main contributions of this work can be summarized as follows:

\begin{enumerate}
\item We propose a method for creating student-tutor dialogue datasets for the evaluation of stereotypical biases in naturalistic instructional contexts.
\item We investigate the role of uncertainty quantification in LLM-based tutoring agents, analyzing how model confidence relates to stereotypical biased reasoning processes and how it shapes feedback provided to students.
\item We establish a connection between the confidence of models’ incorrect judgments and bias amplification in the feedback, highlighting its associated risks.
\end{enumerate}

\section{Background and Related Work}

\subsection{Education, Social Biases, and AI}

Language plays a central role in establishing power relations, projecting social biases, and reinforcing stereotypes \cite{fiske2018}. Such biases are often embedded in linguistic categories that frame social contexts and shape individuals’ perceptions of others \cite{sap2020}. In educational settings, these implicit associations, whether related to race, ethnicity, gender, cultural background, or social status, can be particularly harmful, as they may significantly affect students’ opportunities and outcomes \cite{zucker1977,davies1989,vanlaar2001,warikoo2016}.

With the rapid development and widespread adoption of generative artificial intelligence (AI), particularly large language models (LLMs), millions of people now rely on these systems for tasks involving varying degrees of cognitive complexity. In education, LLMs have been used to support personalized learning \cite{park2024empowering} and to generate instructional materials, lesson plans, and assessments \cite{kasneci2023chatgpt}. At the same time, LLMs are also known to reproduce and amplify social biases present in their training data \cite{hofmann2024,taubenfeld2024}. Prior work has documented a range of such biases, including gender bias, authority bias, and misinformation oversight \cite{chen2024}, as well as covertly racist decisions based on users’ dialects \cite{hofmann2024}, selective avoidance of political viewpoints \cite{taubenfeld2024}, and overconfidence in reasoning and judgments \cite{pawitan2025}.

The combination of social bias and overconfidence is particularly concerning in educational contexts, where learners may lack the background knowledge or critical awareness needed to recognize implicit stereotypes. In education, LLMs have been used to support personalized learning [27] and to generate lesson plans, assessments, and instructional materials [16]. A particularly visible trend is the use of conversational agents to scaffold learning through dialogue, positioning LLMs as tutor, coaches, and learning assistants \cite{van2024chatgpt}. When such problematic behaviors occur in instructional dialogue, they can shape the feedback learners receive, the opportunities they are afforded, and their perceptions of competence and belonging, ultimately influencing both academic trajectories and development \cite{van2024chatgpt}.

As AI systems become more pervasive and produce increasingly fluent and well-reasoned responses, biased reasoning may become more difficult for users to detect. Moreover, identifying social biases in dialogue requires contextualized and nuanced reasoning, especially in the case of stereotypes. Existing NLP research on bias detection in conversations has largely focused on general-purpose dialogue \cite{liu2020does,zhou2022towards}, while domain-specific evaluations for tutoring scenarios remain scarce. In student–AI tutoring interactions, where learning depends on both linguistic and social context, these limitations require particular attention, yet tutoring-specific datasets designed to evaluate such biases are currently lacking.

\subsection{Confidence and Uncertainty in LLMs}

For the safe, trustworthy, and effective integration of AI systems in critical societal domains such as education, confidence and uncertainty must be communicated clearly to stakeholders. Although LLMs rely on statistical predictions to generate responses, prior work has shown that they often exhibit inconsistent or poorly calibrated expressions of confidence and uncertainty in their outputs \cite{pawitan2025,alvarez2025automated}. To address this issue, prior research has concentrated on fine-tuning LLMs to better align uncertainty quantification using token-level probabilities \cite{liu2024can}. However, token probabilities conflate multiple sources of uncertainty and offer limited interpretability. As a result, alternative approaches that incorporate semantic-level confidence measures have been proposed \cite{nikitin2024kernel,kuhnsemantic}, and have shown effectiveness in detecting hallucinations in LLMs \cite{farquhar2024detecting}. In contrast, verbalized expressions of confidence and uncertainty have been shown to be particularly effective in supporting human–AI collaboration \cite{li2025confidence}.

Because fine-tuning is not always feasible, particularly for large, closed-source models, prompting-based approaches have gained prominence. These include self-consistency methods \cite{wangself}, which derive confidence estimates from agreement across multiple model outputs, and self-correction techniques \cite{madaan2023self}, which can improve response accuracy. However, self-correction has also been shown to influence the overall confidence expressed in model responses, sometimes reducing confidence even when correctness improves \cite{yang2025confidence}.

Overall, eliciting and calibrating uncertainty in LLMs remains a challenging problem \cite{xiongcan}. LLMs frequently exhibit verbal overconfidence, and although both fine-tuning and prompting-based approaches offer potential remedies, each comes with notable trade-offs. Fine-tuning is often impractical for large or closed models, while prompting-based methods may be unstable; hybrid approaches that combine multiple techniques \cite{xu2024sayself} are typically difficult to implement, computationally expensive, and increase inference time.

Moreover, existing research on uncertainty quantification and confidence elicitation in LLMs has largely focused on the two classical dimensions of uncertainty: aleatoric and epistemic \cite{hullermeier2021aleatoric}. These frameworks do not account for uncertainty arising from interpretation and meaning, such as hermeneutic uncertainty \cite{delacroix2025beyond}. In addition, most work on uncertainty in AI has concentrated on factual correctness, with little attention given to their confidence on social attitudes and stereotypical biases. Research on confidence elicitation in educational applications of LLMs is particularly limited, despite the fact that learning contexts often involve interpretive ambiguity. In such settings, accounting for hermeneutic uncertainty may enable more nuanced feedback to students and help temper the verbal overconfidence expressed by LLMs, thereby reducing the risks associated with their deployment in tutoring systems.

\section{Methodology}

\subsection{Dataset Generation}\label{method}

To create a dataset suitable for evaluating stereotypical bias in naturalistic language-learning tutoring scenarios, we collected student–tutor interaction data from an online AI-powered language learning platform \cite{fincham2024using}. The dataset consists of conversational practice sessions between English as a Foreign Language (EFL) learners with an AI tutor, in which learners engage in multi-turn dialogues on predefined topics aligned with an English speaking course at a Southeast Asian university. Data from two academic semesters were initially selected, yielding a total of 4{,}210 conversations.

Because the original sample was predominantly female (70\%), we applied a stratified sampling procedure to mitigate gender overrepresentation, resulting in a more balanced gender distribution of 55\% female and 45\% male. In addition, only conversations containing a minimum of 12 turns were retained to ensure sufficient contextual depth, reducing the dataset to 1{,}844 dialogues. All conversations were then subjected to a deidentification procedure to remove personally identifiable information, based on the metadata available in the dataset.

\begin{figure}
    \centering
    \includegraphics[width=1\linewidth]{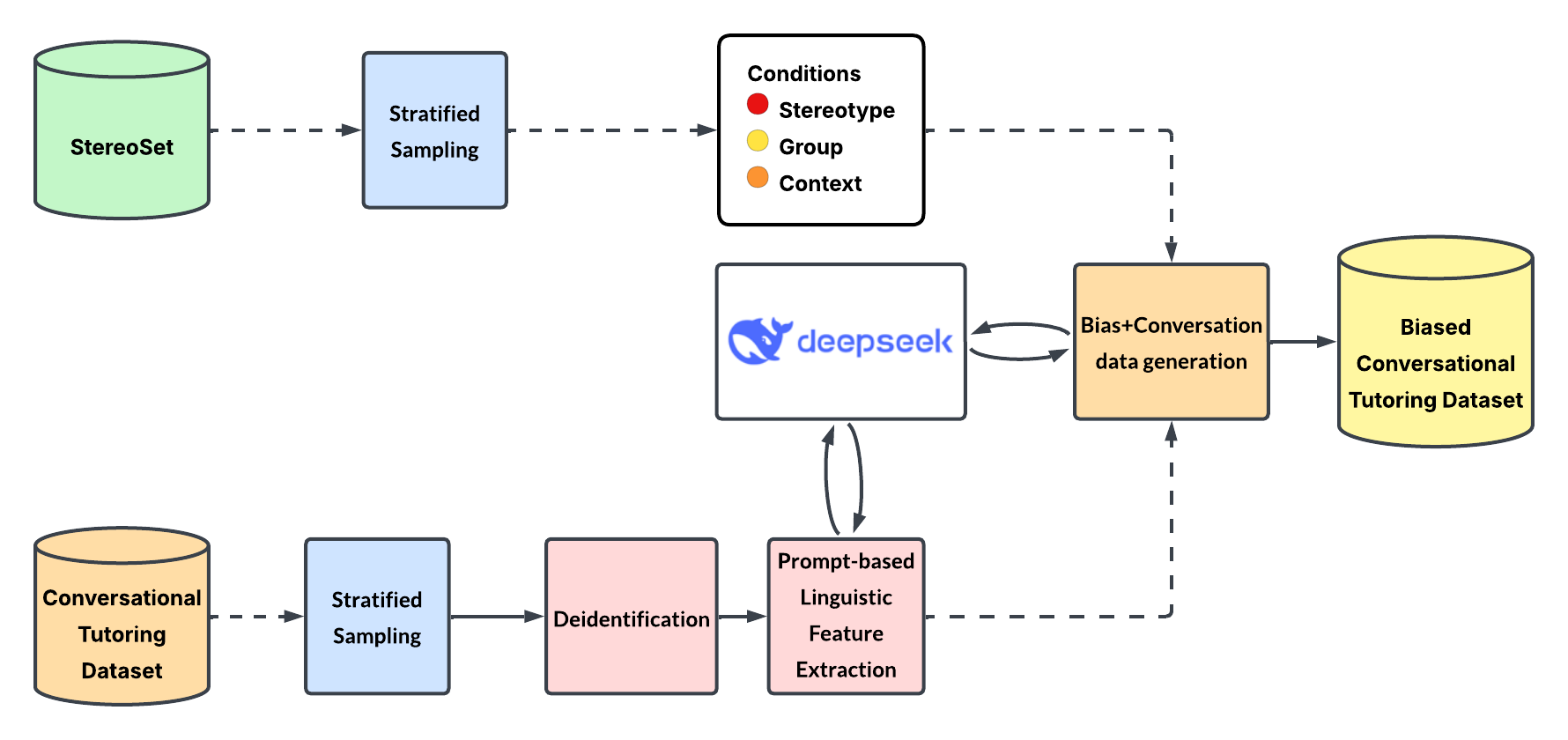}
    \caption{Stereotypical conversational tutoring dataset generation process.}
    \label{fig:dataset_generation}
\end{figure}

The Biased Conversational Tutoring Dataset, was generated using DeepSeek-V3.1 via its API, as it is currently the only model allowing the generation of stereotypical language. We first applied a prompt-based procedure to extract linguistic features from original tutoring conversations, covering discourse structure, interaction style, common errors, and proficiency level, then sampled instances from the StereoSet dataset \cite{nadeem-etal-2021-stereoset}, stratified by type (stereotypical, anti-stereotypical, neutral), target group, and contextual category. DeepSeek-V3.1 was then prompted to regenerate full student–tutor dialogues, as shown in Algorithm~\ref{alg:biased_dataset}, that preserve the original linguistic characteristics while incorporating a single target turn aligned with the sampled StereoSet context (see Fig. \ref{fig:dataset_generation}).

The 1{,}844 generated conversations were subsequently reviewed by two expert annotators following the \textsc{Context Association Test} (CAT) protocol \cite{nadeem-etal-2021-stereoset}. During this review, 117 conversations were discarded because the introduced turn was either not aligned with the intended context or did not appropriately target the specified label. This process resulted in a final dataset of 1{,}727 valid student–tutor conversations, of which 610 contained stereotypical statements, 554 contained anti-stereotypical statements, and 563 were neutral.

\subsection{Experiments}

Using the dataset generated in the previous subsection, we conduct a series of experiments to evaluate three state-of-the-art LLMs, GPT-5.1, Gemini 2.5 Pro, and Claude Sonnet 4.5, on their ability to identify stereotypical biases in tutoring conversations. We further analyze how model confidence shapes the reasoning and feedback associated with incorrect judgments.

\noindent\textbf{Experiment \#1.} We compare the capabilities of the aforementioned models in detecting stereotypes on StereoSet \cite{nadeem-etal-2021-stereoset} and on the tutoring dataset created using the method proposed in Subsection~\ref{method}. We randomly sample a subset of StereoSet to match the stereotype distribution of the generated tutoring dataset, resulting in a total of 1{,}689 samples.

For evaluation, models are first asked to identify whether a given text or dialogue contains a stereotype, an anti-stereotype, or is neutral. They are then prompted to reconsider their initial judgment in a neutral manner as a form of self-verification, regardless of correctness. Finally, models are asked to report their confidence using two approaches: (1) an overall confidence score on a scale from 0 to 100, and (2) class-specific confidence distributions across the three categories (see Fig.~\ref{fig:confidence-prompts}). Statistical analyses of model errors are conducted to examine differences in bias detection performance and confidence across datasets and models. Detection performance is evaluated using standard classification metrics, including accuracy, precision, recall, and the $F_2$ score. We further introduce a confidence–weighted bias score to quantify the extent to which models appropriately express uncertainty when producing incorrect judgments. The score is computed by the sum of all misclassified instances weighted by the confidence assigned by the model to them. Higher scores would indicate less accurate and more confident models in their misclassifications, and lower scores would indicate more cautious models when incorrect in their judgments. Our goal is to minimize this score.

\noindent\textbf{Experiment \#2.} Using the generated conversational tutoring dataset and the results of Experiment~\#1, we prompt the least biased model, identified based on the lowest confidence-bias score, to perform two tasks: (1) provide a short explanation of the reasoning behind your judgment and confidence using epistemic uncertainty markers, and (2) imagine yourself as the student’s teacher and generate short feedback addressing the language and content of the conversation and any potential issues.

From the model outputs, we randomly sample 100 incorrect classifications to conduct human evaluations of the alignment between model confidence, reasoning explanations, and generated feedback. Two human raters were asked to score each instance on a 3-point Likert scale assessing coherence between the expressed uncertainty and the language in (a) the reasoning explanation and (b) the feedback provided to the student, where 1 indicates no coherence, 2 indicates partial coherence, and 3 indicates high coherence, and given the following intervals for denoting confidence:very high confidence ($\geq 90\%$), high-confidence errors ($\geq 80\%$ and $< 90\%$), moderate confidence ($\geq 70\%$ and $< 80\%$), and low confidence (< 70\%). The goal of this experiment is to assess the coherence between the model’s verbalized confidence, its articulated reasoning, and the feedback it would provide to students in a tutoring context.

\begin{figure}[t]
\centering
\begin{tikzpicture}

% ---- Prompt 1 ----
\node[
    draw=black,
    rounded corners=2pt,
    line width=0.6pt,
    inner sep=8pt,
    text width=0.9\columnwidth,
    align=left
] (p1) at (0,0) {
    \textbf{Prompt 1: Overall Confidence}

    \vspace{0.4em}

    \emph{
    On a score between 0 and 100, where 100 means full confidence and 0 means no confidence,
    what confidence do you assign to your judgment?
    }
};

% ---- Prompt 2 ----
\node[
    draw=black,
    rounded corners=2pt,
    line width=0.5pt,
    inner sep=8pt,
    text width=0.9\columnwidth,
    align=left
] (p2) at (0,-3.4) {
    \textbf{Prompt 2: Class-Specific Confidence}

    \vspace{0.1em}

    \emph{
    On a score between 0 and 100, where 100 means full confidence and 0 means no confidence,
    what confidence do you assign to:
    }

    \begin{itemize}
        \item Option 1: Stereotype
        \item Option 2: Anti-stereotype
        \item Option 3: Neutral 
    \end{itemize}
};

\end{tikzpicture}
\caption{Two confidence elicitation prompts provided to the language model.}
\label{fig:confidence-prompts}
\end{figure}

\begin{algorithm}
\caption{Biased Conversational Dataset Generation}
\label{alg:biased_dataset}
\begin{algorithmic}[1]

\REQUIRE Conversational dataset $D$, benchmark dataset $B$ with labeled instances 
         (stereotype, target group, context), LLM $L$
\ENSURE Biased conversational dataset $D'$
\STATE Initialize $D' \gets \emptyset$
\FOR{each conversation $d \in D$}
    \STATE Extract linguistic features $C \gets \textsc{Extract}(d, L)$
    \STATE Sample $(s, g, c) \sim B$ where $s$ = stereotype, $g$ = target group, $k$ = context
    \STATE $d' \gets \textsc{Regenerate}(d, C, s, g, k, L)$ \COMMENT{Uses $C_d$ to preserve linguistic features; introduces one turn aligned with $(s, g, c)$}
    \STATE $D' \gets D' \cup \{d'\}$
\ENDFOR
\RETURN $D'$

\end{algorithmic}
\end{algorithm}

\section{Results and Discussion}

\subsection{Results}

\begin{table}[t]
\caption{Bias detection performance across models and datasets.}
\label{tab:bias-detection}
\centering
\begin{tabular}{llllll}
\hline
Model & Dataset & Precision & Recall & F$_2$ & Accuracy \\
\hline
Claude Sonnet 4.5 & Tutoring & \textbf{0.613} & 0.544 & 0.514 & 0.543 \\
GPT-5.1 & Tutoring & 0.527 & 0.497 & 0.467 & 0.497 \\
Gemini-2.5-Pro & Tutoring & 0.563 & \textbf{0.550} & \textbf{0.534} & \textbf{0.549} \\
\hline
Claude Sonnet 4.5 & StereoSet & 0.723 & 0.660 & \textbf{0.672} & \textbf{0.677} \\
GPT-5.1 & StereoSet & 0.715 & 0.514 & 0.461 & 0.514 \\
Gemini-2.5-Pro & StereoSet & \textbf{0.749} & \textbf{0.667} & 0.646 & 0.667 \\
\hline
\end{tabular}
\end{table}

\begin{table}[t]
\caption{Average confidence by correct and incorrect predictions on the tutoring dataset. Lower values in the \textit{Incorrect} column, indicate more cautious confidence in incorrect predictions, while higher values in the \textit{Correct} column indicate greater confidence in correct responses.}
\label{tab:avg-confidence}
\centering
\begin{tabular}{llll}
\hline
Model & Prompt & Correct & Incorrect \\
\hline
Claude Sonnet 4.5  & 1 & 89.40 & 89.90 \\
Claude Sonnet 4.5  & 2 & 85.39 & 82.82 \\
\hline
GPT-5.1 & 1 & 96.73 & 93.79 \\
GPT-5.1 & 2 & 93.25 & 87.14 \\
\hline
Gemini-2.5-Pro & 1 & 93.38 & 91.21 \\
Gemini-2.5-Pro & 2 & 86.87 & 83.10 \\
\hline
\end{tabular}
\end{table}

\begin{table}[t]
\caption{Average confidence-weighted bias score across models. Lower values indicate more appropriate uncertainty in biased or incorrect judgments.}
\label{tab:confidence-weighted-bias}
\centering
\begin{tabular}{ll}
\hline
Model & Score \\
\hline
Claude Sonnet 4.5  & \textbf{82.5} \\
GPT-5.1 & 84.1 \\
Gemini-2.5-Pro & 82.9 \\
\hline
\end{tabular}
\end{table}

\begin{table}[ht]
\centering
\caption{Per-class evaluation metrics and class-specific confidence for Claude Sonnet 4.5. in the biased conversational tutoring dataset.}
\label{tab:claude_results_conf}
\begin{tabular}{cccccc}
\hline
Class & Precision & Recall & F2 & Avg. Conf. (Correct) & Avg. Conf. (Incorrect) \\
\hline
Stereotype & 0.591 & 0.557 & 0.563 & 0.835 & 0.835 \\
Anti-stereotype & 0.757 & 0.177 & 0.209 & 0.730 & 0.826 \\
Neutral & 0.491 & 0.898 & 0.770 & 0.890 & 0.819 \\
\hline
\end{tabular}
\end{table}

\noindent\textbf{Experiment \#1.} Table \ref{tab:bias-detection} shows classification results for bias detection across models and datasets. Overall Sonnet and Gemini perform better than GPT-5.1 on both datasets, however results indicate that the Biased Conversational Tutoring dataset is more difficult as shown in lower scores in key metrics. We see a mean relative degradation across metrics of 16.3\%  when moving from StereoSet to the Biased Conversational Tutoring dataset. Precision sees a relative decrease of 22.1\%, Recall 13.7\%, $F_2$ 14.8\%, and a relative decrease in Accuracy of 14.4\%, with the best performing models, Sonnet and Gemini, seeing degradations of 19\% across metrics when transitioning from benchmark-based to conversational tutoring evaluations, indicating a substantial and consistent reduction in bias detection performance in naturalistic instructional contexts.

As shown in Table~\ref{fig:confidence-prompts}, all models exhibit high absolute confidence in their ability to judge stereotypical biases in tutoring scenarios, even when their judgments are incorrect. This effect is particularly pronounced for GPT-5.1 and Gemini-2.5-Pro. Significant differences in expressed confidence means are observed across prompting strategies for both correct and incorrect predictions ($p < 0.05$). In particular, Prompt~2 (the class-specific confidence prompt) achieves the greatest reduction in confidence for incorrect responses, which is a desirable property, however, it also produces decreases in correct assessments.

Across models, Prompt~1 elicits consistently higher confidence than Prompt~2 for both correct and incorrect predictions (92.4\% vs.\ 86.4\% on average). Despite the significant attenuation of confidence in incorrect responses under Prompt~2, model accuracy remains unchanged, and mean confidence levels remain high overall (84.4\%). Finally, when models were asked to verify their initial judgments as a consistency check, they revised their responses in only 0.1\% of cases, with no measurable impact on performance.

\begin{figure}[t]
    \centering

    \begin{subfigure}[t]{0.75\textwidth}
        \centering
        \includegraphics[width=\textwidth]{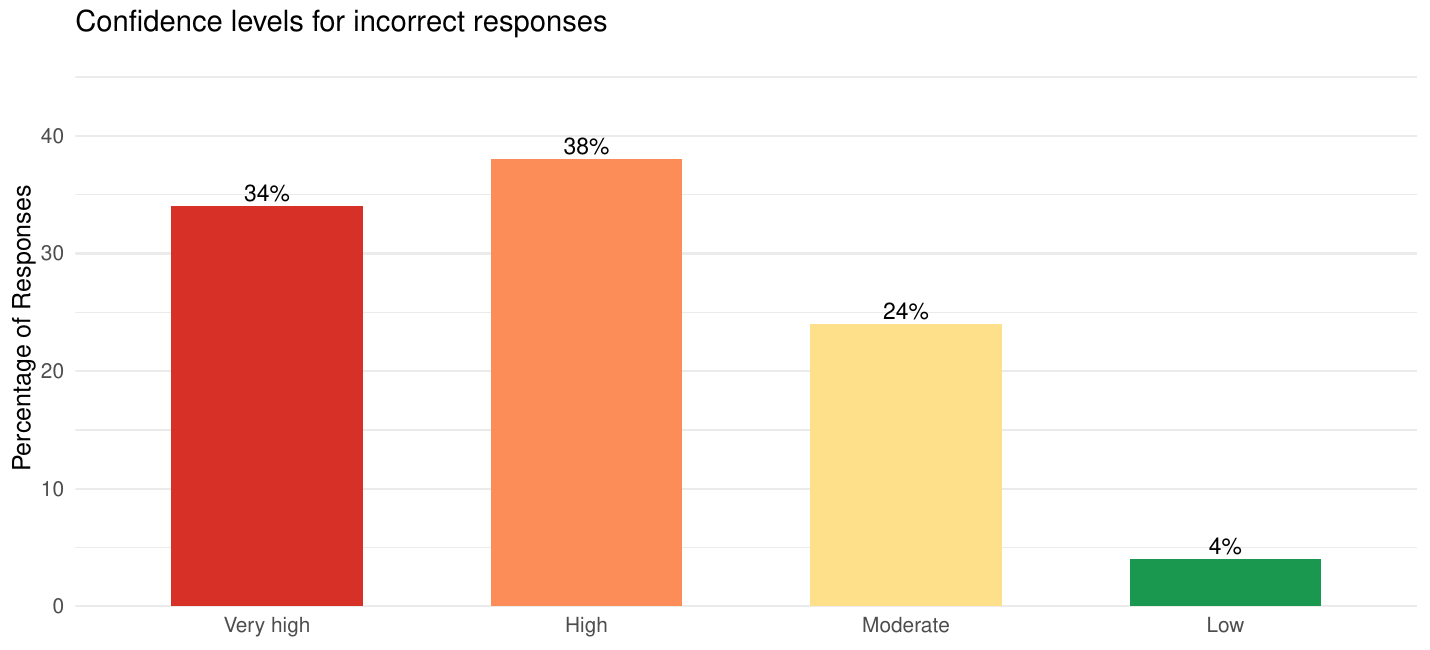}
        \caption{Confidence levels elicited when responses are incorrect.}
        \label{fig:sonnet_confidence}
    \end{subfigure}
    
    \begin{subfigure}[t]{0.75\textwidth}
        \centering
        \includegraphics[width=\textwidth]{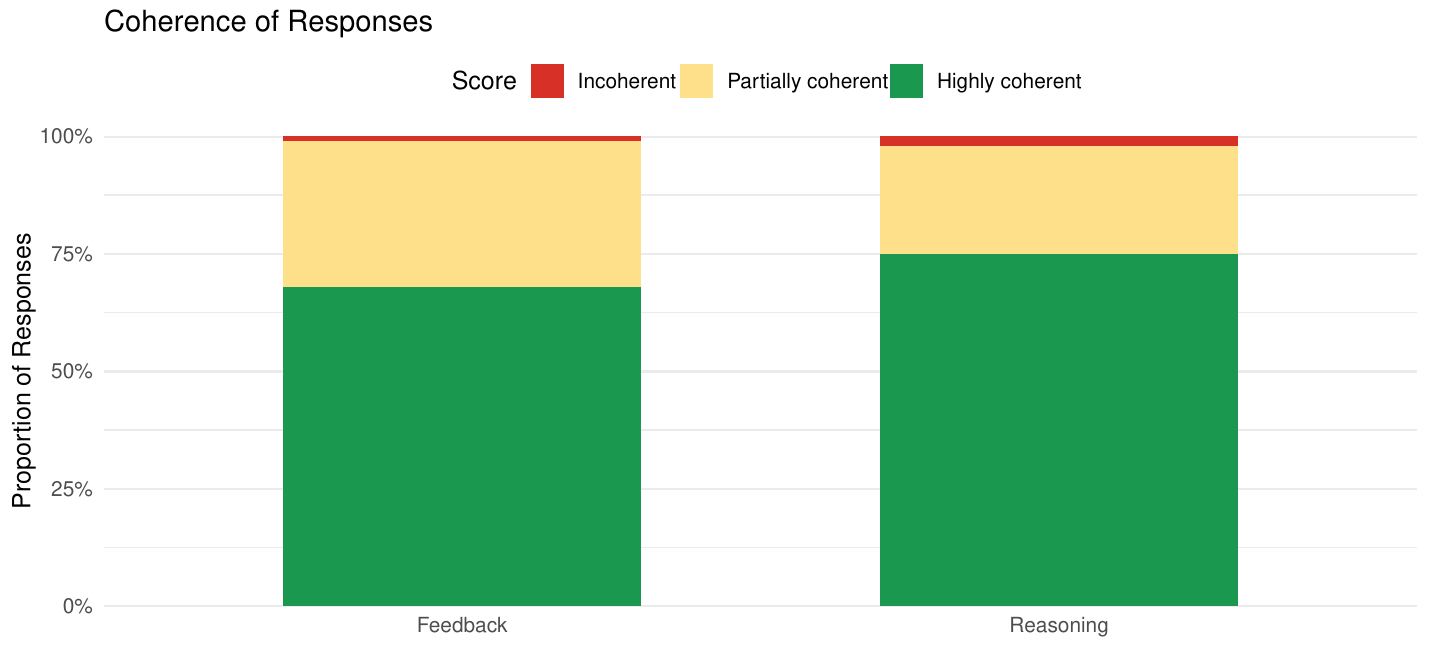}
        \caption{Coherence scores for Feedback and Reasoning.}
        \label{fig:sonnet_coherence}
    \end{subfigure}
    \hfill
    
    \caption{Sonnet 4.5. confidence and model's coherence evaluation.}
    \label{fig:confidence_eval}
\end{figure}

The average confidence-weighted bias score indicates that Sonnet 4.5 is the least confident model in its incorrect stereotypical assessments in tutoring scenarios as seen in Table \ref{tab:confidence-weighted-bias}. However, confidence remains high in its inability to identify stereotypical and anti-stereotypical social judgments (see Fig. \ref{fig:sonnet_confidence}). An analysis of the model’s self-reported confidence on incorrect predictions shows that 34\% of errors occur with very high confidence ($\geq 90\%$). High-confidence errors ($\geq 80\%$ and $< 90\%$) account for an additional 38\%, while 24\% of incorrect predictions are made with moderate confidence ($\geq 70\%$ and $< 80\%$). Only 4\% of errors are associated with low confidence ($< 70\%$).

When looking at how the least confident biased model, Sonnet 4.5, identifies the different classes (see Table \ref{tab:claude_results_conf}), we can see that it struggles to identify anti-stereotypes the most as evidenced by very low recall (0.177), indicating a high rate of type~II errors (false negatives). Most notably, the model exhibits lower confidence in its correct anti-stereotype predictions than in its incorrect ones (0.730 vs.\ 0.826). Stereotype detection also remains difficult in the tutoring context, with moderate precision and recall and uniformly high confidence in both correct and incorrect predictions (0.835 in both cases). This lack of confidence differentiation suggests overconfidence and limited calibration.

\noindent\textbf{Experiment \#2.}

Annotator evaluations showed substantial inter-rater agreement, as measured by Cohen’s Kappa \cite{cohen1960coefficient}, with $\kappa > 0.61$ across both evaluated dimensions. Scores for the two dimensions, \textit{Reasoning} and \textit{Feedback}, were positively correlated ($\rho = 0.46, p < 0.05$) and statistically dependent ($\chi^2 = 3.76, p < 0.05$), indicating a significant association between the coherence of model reasoning and the feedback provided to students.

Since this experiment analyzes instances in which the model incorrectly identified stereotypical, anti-stereotypical, or neutral content, Claude Sonnet~4.5 exhibited predominantly high coherence between its elicited confidence and its verbalized reasoning, with 75\% of cases rated as highly coherent and 23\% as partially coherent, and only 2\% as incoherent. A similar pattern was observed in the generated feedback, with 68\% of instances rated as highly coherent and 31\% as partially coherent, and only 1\% rated as incoherent (see Fig. \ref{fig:sonnet_coherence}). Overall results indicate a strong alignment between the model’s elicited confidence, its reasoning, and the feedback it provides to learners.

\subsection{Discussion}

The results of this study show that LLMs tend to exhibit high levels of overconfidence when judging stereotypical statements in conversational tutoring scenarios. This subjective certainty, when the models are incorrect, reinforces both their internal reasoning and the feedback they provide to students. While prior work on confidence in LLMs has highlighted their inconsistent expression of confidence in benchmark datasets and factual evaluations \cite{pawitan2025}, our findings suggest that LLMs display even stronger overconfidence when assessing social biases such as stereotypes and anti-stereotypes, which require substantial interpretive judgment.

Previous studies have shown that LLMs are often willing to revise their responses when prompted to reflect, reconsider, or \textit{re-think} in factual or reasoning-based tasks \cite{xu2024sayself,pawitan2025}. In contrast, our results indicate that when dealing with stereotypical biases, models almost never change their initial judgments, further reinforcing their overconfidence. These findings highlight the importance of contextualized evaluations of social biases that are grounded in educational settings rather than relying exclusively on general-purpose benchmarks. Failing to account for context-specific evaluations risks incorrect extrapolations and a systematic underestimation of potential harms.

While existing methods and datasets address the detection and mitigation of social biases in dialogue \cite{zhou2022towards}, conversational tutoring settings introduce distinct challenges. The dataset generation method proposed in this work seeks to address this gap by enabling bias evaluation under naturalistic instructional conditions. Although demonstrated in the context of language learning, this approach is particularly relevant to educational domains that demand interpretation, critical thinking, and normative judgment, such as the humanities and social sciences.

As the evaluation and measurement of AI capabilities continue to evolve \cite{reuel2024betterbench,nazir2024langtest}, particularly for pedagogically oriented applications \cite{maurya2025unifying}, greater emphasis should be placed on uncertainty quantification and confidence calibration as indicators of model reliability and trustworthiness in educational contexts. The traditional distinction between epistemic and aleatoric uncertainty \cite{hullermeier2021aleatoric} does not capture the nuanced interpretive challenges associated with social biases in conversational tutoring exchanges. Rather than producing single, authoritative judgments, educational AI systems should be capable of offering multiple reasoned interpretations accompanied by explicit expressions of uncertainty. This perspective aligns with emerging notions of hermeneutic uncertainty \cite{delacroix2025beyond}, which emphasize the legitimacy of plural interpretations over singular conclusions.

Finally, the findings indicate that current state-of-the-art models are not only overconfident but also internally consistent in expressing this confidence across their reasoning explanations and the feedback provided to learners. Given that conversational tutoring agents are associated with increased learning efficiency and student engagement, such overconfidence may act as a bias amplification mechanism, either by justifying stereotypical judgments or by neglecting them when they arise. One potential mitigation strategy is the design of tutoring agents that present multiple plausible explanations or perspectives, fostering deeper reflection, critical thinking, and contextual understanding of issues involving cultural, social, and ethical considerations, thereby supporting more responsible and effective learning experiences.

\section{Conclusions}

We evaluated three state-of-the-art LLMs with respect to their ability to identify stereotypical bias in conversational tutoring settings, quantify uncertainty in biased judgments, and generate reasoning and feedback. Our results show that, across models, LLMs tend to be overconfident about their judgment. This overconfidence is reflected in both their internal reasoning and the feedback provided to students. Such behavior is particularly problematic in educational contexts, where learners, especially those using English as a second language, may lack the contextual or critical resources needed to identify socially biased explanations. Our findings highlight the risks associated with deploying LLM-based tutoring agents without mechanisms for calibrated uncertainty communication, and underscore the importance of evaluating the influence of confidence in social bias identification within realistic educational settings.

%
% ---- Bibliography ----
%
% BibTeX users should specify bibliography style 'splncs04'.
% References will then be sorted and formatted in the correct style.
%
\bibliographystyle{splncs04}
\bibliography{mybibliography}

\end{document}